# Discovering Interpretable Machine Learning Models in Parallel Coordinates

Boris Kovalerchuk
*Dept. of Computer Science*
*Central Washington University,*
USA
BorisK@cwu.edu

Dustin Hayes
*Dept. of Computer Science*
*Central Washington University*
USA
Dustin.Hayes@cwu.edu

*Abstract*—This paper contributes to interpretable machine learning via visual knowledge discovery in parallel coordinates. The concepts of hypercubes and hyper-blocks are used as easily understandable by end-users in the visual form in parallel coordinates. The Hyper algorithm for classification with mixed and pure hyper-blocks (HBs) is proposed to discover hyper-blocks interactively and automatically in individual, multiple, overlapping, and non-overlapping setting. The combination of hyper-blocks with linguistic description of visual patterns is presented too. It is shown that Hyper models generalize decision trees. The Hyper algorithm was tested on the benchmark data from UCI ML repository. It allowed discovering pure and mixed HBs with all data and then with 10-fold cross validation. The links between hyper-blocks, dimension reduction and visualization are established. Major benefits of hyper-block technology and the Hyper algorithm are in their ability to discover and observe hyper-blocks by end-users including side by side visualizations making patterns visible for all classes. Another advantage of sets of HBs relative to the decision trees is the ability to avoid both data overgeneralization and overfitting.

## I. Introduction

Further progress in Machine Learning (ML) heavily depends on acceptance and interpretability of produced models. Visual knowledge discovery is a promising avenue to deal with this challenge [Kovalerchuk et al., 2018-2020]. Below we present a visual knowledge discovery approach based on parallel coordinates [Inselberg, 2009] that include pattern discovery, data and model visualization, dimensionality reduction, and model simplification. This paper is focused on supervised classification models.

Building interpretable ML models often requires putting the end-user in the driver seat of development of a trusted model. The end-users often are not experts in machine learning, but domain experts. The formal ML models often are not comprehensible for them, serving as black boxes. The visual knowledge discovery approach allows uncovering ML models, making them interpretable for the end-users and putting them in the driver seat of model development.

Parallel coordinates are very well suited to visual knowledge discovery [Inselberg, 1998, 2009; Estivill-Castro et al., 2020; Sansen et al, 2017; Xu et al, 2007]. They visualize multidimensional data in two dimensions without loss of any multidimensional information. In addition, parallel coordinates support interpretability, because they use the original attributes which have direct domain meaning for the domain end-users.

The proposed approach is based on the concepts of *hypercubes* (HCs) and *hyper-blocks* (*HBs*) that are naturally visualized in parallel coordinates with *clear interpretation*. They provided the end-users the ability to be active in interactive constructing and improving such models not only using models. Decision tree models with parallel coordinates have been proposed in [Estivill-Castro et al., 2020; Tam et al, 2016]. We show that the hyper-block approach produces models that are more general than decision trees (DTs).

Section II defines: (1) hypercubes and hyper-blocks along with their visualization in parallel coordinates and (2) algorithms to discover them in interactive and automatic ways. We first focus on individual HCs and HBs, then on multiple HBs. With multiple HBs we focus on discovering and visualizing pairs of non-overlapping HBs and combination of HBs. Linguistic description of visual patterns is presented next.

Section III presents supervised learning in parallel coordinates. First, we describe challenges of this learning, then we introduce a classification algorithm Hyper for pure and mixed/impure hyper-blocks. The comparison with decision strees and Hyper models as generalized decision trees follow. To test the approach, we used a benchmark Wisconsin Breast Cancer (WBC) data [Dua, Graff, 2019]. Next, we established a link between hyper-blocks, dimension reduction and visualization of lower dimensional hyper-blocks. Section IV presents conclusion and future studies in machine learning in parallel coordinates. The proposed approach was implemented in VisCanvas 2.0 software, which is based on the previous version VisCanvas [2018].

## II. Visual Discovery of Hyper-Blocks

### A. Hypercubes and hyper-blocks

Hypercubes (HCs) and hyper-blocks (HBs) are interpretable concepts that naturally present in parallel coordinates.

Definition. A **hypercube** (**n-cube**) is a set of n-D points $\{\mathbf{x}=(x_1,x_2,\dots,x_n)\}$ with **center** in n-D point $\mathbf{c}=(c_1,c_2,\dots,c_n)$ and **side length** $L$ such that

$$\forall i \ \| x_i - c_i \| \leq L/2 \qquad (1)$$

A hypercube is an n-dimensional generalization of ordinary square ($n=2$) and cube ($n=3$). A *unit hypercube* has $L=1$. A *binary hypercube* is an n-cube in the binary n-D space $E^n$.

Definition. A **hyper-block (hyperrectangle, n-orthotope)** is a set of n-D points $\{\mathbf{x}=(x_1,x_2,\ldots,x_n)\}$ with **center** in n-D point $\mathbf{c}=(c_1,c_2,\ldots,c_n)$ and **side lengths** $\mathbf{L}=(L_1, L_2,\ldots, L_n)$ such that

$$\forall i \ \|x_i - c_i\| \leq L_i/2 \qquad (2)$$

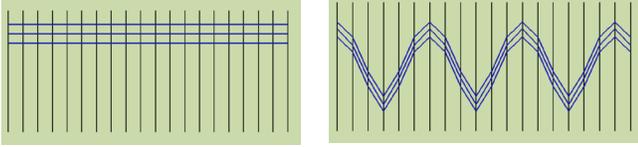

(a) hypercube A with equal coordinates  (b) zigzag hypercube B

Fig. 1. Examples of hypercubes.

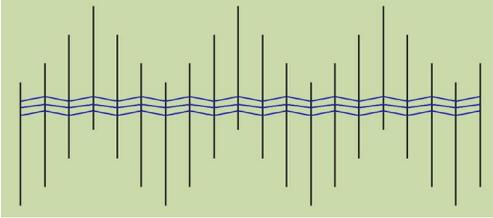

Fig. 2. Hypercubes in adjustable parallel coordinates shifted up and down to simplify hypercube patterns.

Thus, the n-cube is the special case of a hyper-block. The concept of a hypercube is illustrated in Figs. 1 and 2 in 20-D parallel coordinates. The middle lines show 20-D centers of these HCs. Fig. 2 shows hypercubes in *adjustable parallel coordinates* (APC), where coordinates are shifted up and down to make hypercube patterns simpler for human perception [Kovalerchuk, Grishin, 2019]. Fig.3 shows all Wisconsin Breast Cancer (WBC) data in parallel coordinates

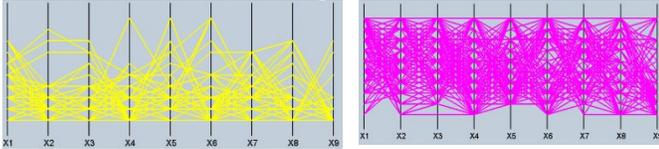

Fig. 3. All cases of two classes of WBC data.

*B. Algorithm to generate hyper-blocks*

Hyper-blocks can be composed of hypercubes that share some dimensions and have their centers within the range of each other. However, not all combinations of HCs are hyper-blocks.

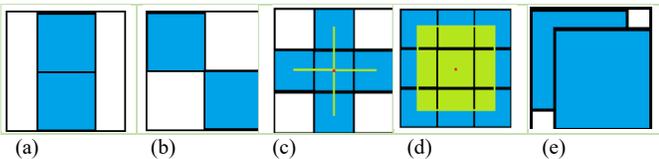

(a)   (b)   (c)   (d)   (e)

Fig. 4. Adjacency options for 2-D hypercubes.

Fig. 4 shows different alternatives how HCs can be combined. In Fig. 4a, two adjacent "2-D hypercubes" (squares) share $n$-1 dimensions (1-D line for the squares) and can form a single hyper-block. In Fig. 4b, two adjacent "2-D hypercubes" (squares) share only $n$-2 dimensions (a single point) and do not form a single hyper-block. Figs. 4cd show more adjacent HCs for the square in the center, while Fig. 4e shows overlapping HCs that do not from a single hyper-block. Due to these complex relations between HCs and HBs, in this work, we start with small pure HCs seeded in the individual n-D points as centers and grow them to HBs keeping them pure.

Below we present the steps of the **Merger Hyper-blocks (MHyper)** algorithm for pure and dominant hyper-blocks.

Step M1: Seed an initial set of **pure** HBs with a single n-D point in each of them (HBs with length equal to 0).
Step M2: Select a HB **x** from the set of all HBs.
Step M3: Start iterating over the remaining HBs. If $HB_i$ has the same class as **x** then attempt to combine $HB_i$ with **x** to get a pure HB.

 Step M3.1: Create a joint HB from $HB_i$ and **x** that is an envelope around $HB_i$ and **x** using the minimum and maximum of each dimension for $HB_i$ and **x.**
 Step M3.2: Check if any other n-D point **y** belongs to the envelop of $HB_i$ and **x.** If **y** belongs to this envelope add **y** to the joint HB.
 Step M3.3: If all points **y** in the joint HB are of the same class then remove **x** and $HB_i$ from the set of HBs that need to be changed.

Step M4: Repeat step 3 for all remaining HBs that need to be changed. The result is a *full pure HB* that cannot be extended with other n-D points and continue to eb pure,
Step M5: Repeat step 2 for n-D points do not belong to already constructed full pure HBs.
Step M6: Define an *impurity threshold* that limits the percentage of n-D points from opposite classes allowed in a **dominant** HB.
Step M7: Select a HB **x** from the set of all HBs.
Step M8: Attempt to combine **x** with remaining HBs.
 Step M8.1. Create a joint HB from $HB_i$ and **x** that is an envelope around $HB_i$ and **x**.
 Step M8.2: Check if any other n-D point **y** belongs to the envelop of $HB_i$ and **x.** If **y** belongs to this envelope add **y** to the joint HB.
 Step M8.3: Compute impurity of the $HB_i$ (the percentage of n-D points from opposite classes introduced by the combination of **x** with $HB_i$.)
 Step M8.3 Find $HB_i$ with lowest impurity. If this lowest impurity is below predefined impurity threshold create a joint HB.
Step M9: Repeat step 7 until all combinations are made.

*C. Individual hyper-blocks*

As was illustrated above a major **challenge** for discovering patterns is in the **overlapping of lines (polylines)** that represent individual cases (n-D points). Hundreds of lines can follow exactly the same path under a given visualization resolution even when the values differ. A user cannot distinguish such lines and cannot see how many lines are in the hypercube and which line segments belong to which n-D points.

**Histograms for individual hyper-blocks**. Histograms help to address this issue. Fig. 5 visualizes a **distribution** of n-D points (pink lines) within a hyper-block as white rectangles and black lines of the same length. Each white rectangle and black line correspond to a data quantile. The width of the pink lines indicates their frequencies similar to [15]. Fig. 5 clearly shows where these 9-D points are distributed in the 9-D WBC space.

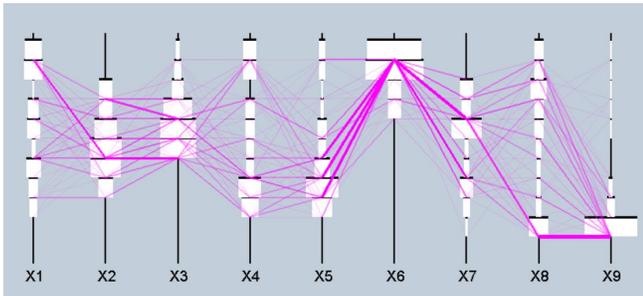
Fig. 5. Distribution of 9-D points in parallel coordinates with white rectangles to show quantiles for two WBC HBs.

Figs. 6 and 7 show a way to represent visually a distribution of polylines in each hyper-block **side-by-side** with quantiles in a separate window.

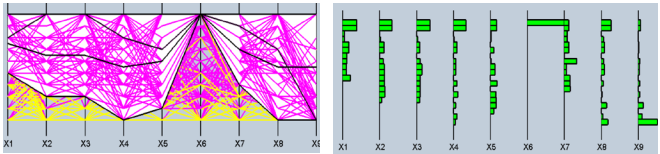

(a) HB A with white background  (b) Frequency of cases for HB A.
Fig. 6: Hyper-block A and its distribution using quantiles.

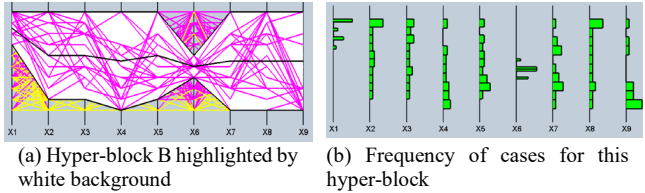

(a) Hyper-block B highlighted by white background  (b) Frequency of cases for this hyper-block
*Fig. 7: Hyper-block B and its distribution using quantiles.*

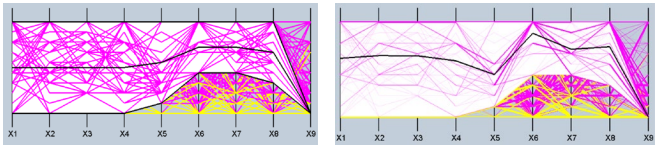

Fig. 8. Normal visualization of a hyper-block.  Fig. 9. Visualization of a hyper-block with line frequencies enbeded.

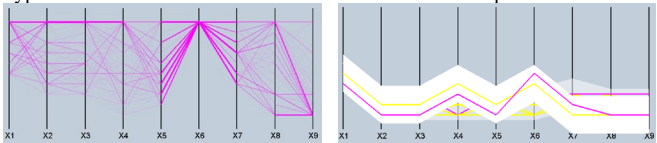

Fig. 10. View of the hyper-block's lines with their frequencies.  Fig. 11. Mixed hyper-blocks.

Also Figs. 6 and 7 clearly show the differences between HBs A and B especially in coordinate X6 where their distributions do not overlap. Figs. 8 and 9 show another example of HBs. The width of lines indicates their frequencies in Fig. 9. Thus, here we have **frequencies embedded** to parallel coordinates. These examples show that discovering such distinct HBs leads to interpretable classification rules. Figs. 10-11 show more HB with **embedded frequency** by the line width. Fig. 10 shows a pure HB with a black line that represents the mean of frequency withing the HB. Fig.11 shows a mixed HB.

D.  Multiple hyper-blocks

**Side-by-side view of hyper-blocks**. Below we consider a challenging task of visualizing hyper-blocks that *do not overlap in n-D space to also not overlap in 2-D space* when drawn in parallel coordinates. This is possible in very simple situations, Figs. 6 and 7 show HBs that do not overlap in 9-D space, but overlap in parallel coordinates in 2-D. To avoid drawing such HBs one on the top of another one, we presented them in separate Figs. 6 and 7, i.e., side-by-side. Fig. 12 shows how this side-by-side option is implemented. Another approach is representing the HBs in a **3-D space**. First, the camera is placed to only view two dimensions normally. Then the user can tilt the camera to reveal the 3-D space where each hyper-block is visible in a separate plane in the z-coordinate.

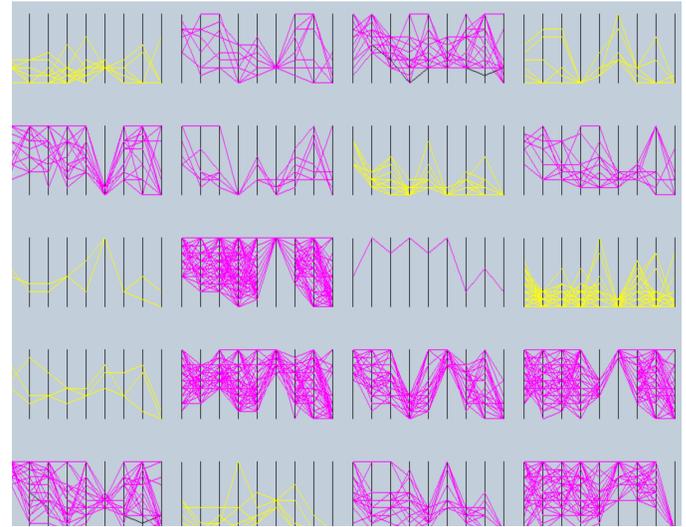
Fig. 12. All WBC discovered pure HBs of two classes side by side.

**Non-overlapping areas**. The next approach is showing the **differences** in distribution of HBs in n-D space of parallel coordinates. First, all unique pairings of hyper-blocks are made. Then a rectangle is drawn on the dimensions where a given pair **do not overlap**. This was visualized in several ways. Fig. 13 shows one of these ways. The more pairs that do not overlap in a coordinate, the darker the rectangle is in Fig. 13. It is clear from Fig. 13 that HBs less overlap in X6 coordinate.

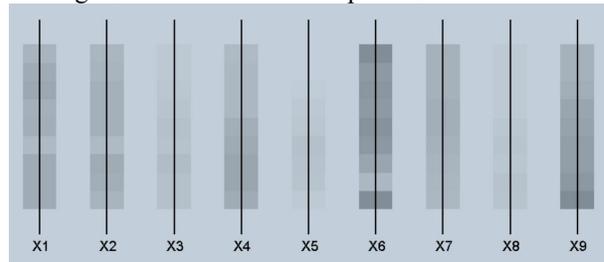
Figure 13. Locations where pairs of hyper-blocks do not overlap.

E.  *Discovering and visualizing pairs of non-overlapping HBs*

The difficulties to recognize visually patterns of different classes increase significantly when polylines of opposing classes heavily overlap requiring computational, and interactive tools to make the difference clearer.

One of the approaches is finding subsets of coordinates where the difference is quite clear. The challenge of the search for such subsets is in multiplicity of HBs that represent each class. An individual pair of HBs from opposing classes can be quite distinct in some parallel coordinates. However, as the number of HBs that represent classes increases, the likelihood of overlap of pairs in each coordinate goes up. It means that the likelihood for gaining valuable insights goes down past just comparing two HBs. Therefore, we search for distinctive pairs of HBs and visualize each such pairs individually.

Below we demonstrate results of the successful search in WBC data that is implemented in VisCanvas 2.0. A sequence of automatic and interactive steps to reach this point taken in VisCanvas are as follows:
1. Auto-generate hyper-blocks for opposing classes,
2. Chose hyper-blocks which contain the most of data n-D points for each class,
3. Auto-reduce dimensions of these hyper-blocks to dimensions where there these HBs do not overlap,
4. Visualize these HBs in reduced dimensions,
5. Ascertain a classification rule based on these HBs and evaluate its accuracy.

These steps resulted in 93.7% accuracy for these WBC data. Fig. 14 shows two such discovered HBs.

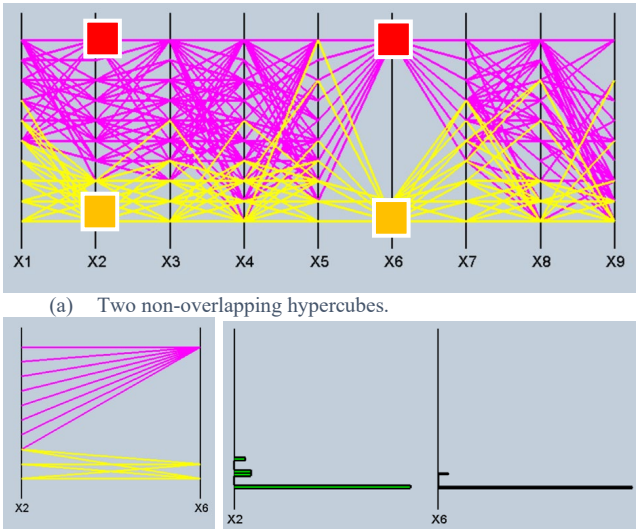

(a) Two non-overlapping hypercubes.

(b HCs from (a) in most informative coordinates X2 and X6 with ffrequency visualization of yellow HB.

Fig. 14. Parallel coordinates with two hyper-blocks that contain the most amount of data for their respective classes.

*F. Exploration and combination of hyper-blocks*

**Exploration of hyper-blocks** in VisCanvas 2.0 starts with selecting a line that corresponds to an n-D point of interest and constructing a hypercube with this line as its center line and a specific distance from this line, e.g., 0.2. The borders of HCs that can be toggled on and off. Fig. 15 shows hypercubes with and without their lines.

**Combination Modes**. We explored three different modes to combine hypercubes together. The first mode (M1) checks whether all dimensions of each center are *within range* of each other, i.e., overlap of these hypercubes. The second mode (M2) checks if *at least a single point* exists within both hypercubes. Both modes to be successful require additional check of presence of cases from other classes that can be in the joint hyper-block. The third mode (M3) checks if *n–1 dimensions* of both hypercubes are *the same* and *one dimension* of each center is within range of the other. This condition rarely happens, and Fig. 16 illustrates it.

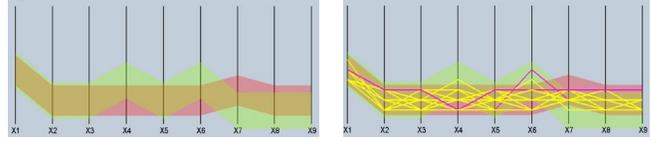

Fig. 15. Hypercubes without lines.    Fig. 16. Hypercubes with lines.

In addition, a threshold display allows showing hypercubes that have a certain number of elements within them. Some hypercubes and hyper-blocks **overlap** since several n-D points exist in many of them. Finding and removing overlaps and defining **joined** new hypercubes or hyper-blocks will minimize the number of them. Hyper-blocks that have a **low purity** ratio are not good candidates to combine with other hyper-blocks. Therefore, it is better to **refuse** these HBs, e.g., by holding them in their own list of refused HBs. These blocks can then be drawn in a different color, toggled on/off, etc. Below we present algorithms and results of computational experiments for combining HBs and VisCanvas 2.0 features that support it.

VisCanvas 2.0 checks that different n-D points produce the same HBs and **eliminates duplicates**. It combines adjacent hypercubes by checking every dimension of the current HC against every dimension of every other HC. If all differences divided by 2 are less than or equal to the threshold value, then they are adjacent. This is beneficial, because it can potentially combine a neighborhood consisting of dozens of cubes into a single hypercube. On the other hand, the hypercubes in this neighborhood may have provided insight into an area of that space that must be discovered manually. Finally, VisCanvas 2.0 counts the total number of unique hypercubes and the number of them that have cases of multiple classes. This is displayed for the user to analyze and to check for overfitting.

A user can view the ratio of HCs that have cases of multiple classes compared to the total number of unique cubes. For WBC data it shows a HC that contains 93 cases from class 1, and two cases from class 2 that are visualized in different colors. Multiple algorithms can be designed to combine HBs. One of them is presented below.

*G. Linguistic description of visual patterns*

**Linguistic descriptions of hyper-blocks combined with distribution visualization.** A linguistic pattern description is designed to simplify capturing the visual patterns by users. Often such descriptions are more concise and "natural" for humans. It takes into consideration the entire n-dimensional space and not just the sub-space that a given hyper-block occupies. A user can activate a linguistic description by using a simple UI. It produces a pop-up window that shows a short paragraph describing the current visualization in a linguistic form. In this way a plainer English explanation is available. Currently, these descriptions identify data concentration in the lower, middle, and upper thirds of the coordinates, while more detailed splits are under development. The current implementation allows merging descriptions of dimensions that are concentrated in the same thirds of them into a compact single description as Fig. 17a shows.

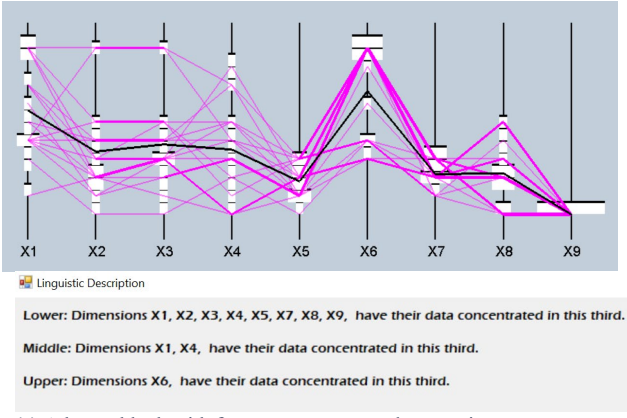

(a) A hyper-block with frequency pattern and annotation.

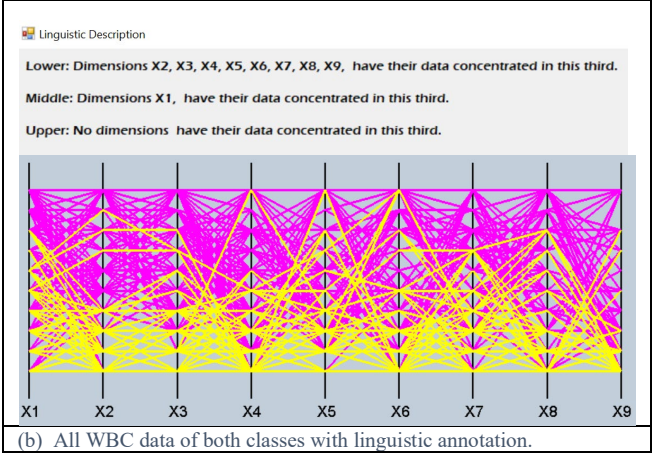

(b) All WBC data of both classes with linguistic annotation.

Fig. 17. WBC data visualized with linguistically annotated patterns.

Due to occlusion, the distribution of n-D points can be corrupted in visualization as it is the case in Fig. 17b. A linguistic description avoids such corruption because it is based on the statistical data analysis. It shows advantages of combining visualization, frequencies, and linguistic description. Fig 17a shows all three aspects of the hyper-block: n-D points, their frequency distribution and linguistic description. While linguistic description in Fig. 17 is well structured relative to three areas of data concentration (lower, middle, upper) in the respective coordinates, it is not in the form of traditional natural language (NL) sentences. Fig. 18 shows modified linguistic descriptions as traditional NL sentences. It is also more concise and allows the user to see the differences between classes in a linguistic form. See Fig. 18.

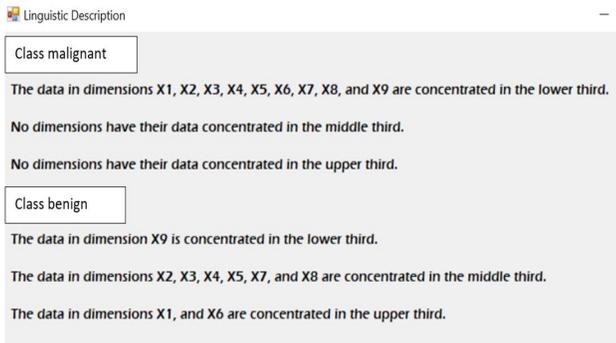

Fig. 18. Linguistic description of all WBC data.

### III. SUPERVISED LEARNING IN PARALLEL COORDINATES

#### A. Hyper-block-based classification algorithm (HYPER)

##### A.1. Mixed and pure hyper-blocks

A **hyper-block-based classification (Hyper) algorithm** is a supervised learning algorithm that learns and visualize interpretable rules in parallel coordinates that is implemented in VisCanvas 2.0 software. The main idea of Hyper algorithm is using n-D hyper-blocks for classification of n-D points (cases/samples). First it searches for *dominant hyper-blocks* where most of n-D points belong to a single class. Say, $HB_1$ is dominant for class 1, $HB_2$ is dominant for class 2 and so on. If a new n-D point **x** *belongs* to a respective HB where class $C_i$ dominates then **x** is classified to class $C_i$. If **x** does not belong to any dominant HB, then **x is** classified to the *nearest HB*. If there are several competing HBs nearby, then the those nearest $k$ HBs vote and **x** is classified to the class with *majority votes*. If there is not enough HBs to vote in the vicinity, the Hyper algorithm refuses to classify this n-D point **x**.

The novelty and benefits of Hyper algorithm are in integration with visualization in parallel coordinates and respective visual knowledge discovery. Its basic algorithmic ideas can be traced back to k-NN first developed in 1951 and the algorithm of computation of estimates [Zhuravlev, Nukiforov, 1971]. There are also recent works on hypercubes relevant to k-NN [Qin et al, 2020; Ohmaind et al, 2020].

Below we introduce the notation and describe the Hyper algorithm in more details. Let $HB_i$ be a hyper-block where n-D points of *class $C_i$ dominate*. Also let $\{HB\}_k$ be a set of *k nearest HBs* for n-D point **x**, which *vote for class* $C_i$, i.e., *most* of these hyper-blocks are class $C_i$ dominant HBs.

The Hyper algorithm learns rules in the following forms:

Rule 1: If n-D point **x** belongs to any dominant hyper-block $HB_i$, for class $C_i$, then **x** belongs to class $C_i$.

Rule 2: If $HB_i$ is a nearest hyper-block for n-D point **x,** then **x** belongs to class $C_i$.

Rule 3: If $\{HB\}_k$ is a set of *k* nearest hyper-blocks for n-D point **x** and the majority of $\{HB\}_k$ vote for class $C_i$, then **x** belongs to class $C_i$.

More formally these rules are:

$$R_1: \exists\, HB_i\, (\mathbf{x} \in HB_i) \Rightarrow \mathbf{x} \in C_i. \quad (1)$$

$$R_2: (HB_i \text{ is a nearest HB for } \mathbf{x}) \Rightarrow \mathbf{x} \in C_i \quad (2)$$

$$R_3: (\{HB\}_k \text{ is } k\text{-}NN \text{ HBs for } \mathbf{x}) \& (\text{Vote}(\{HB\}_k)=C_i) \Rightarrow \mathbf{x} \in C_i \quad (3)$$

The **major steps of Hyper algorithm** to learn $R_1$-$R_3$ are:

**Step H1**: Split data to *training data $T_r$* and *validation data $V_d$* (e.g., by 10-fold cross validation) and split $Tr$ to $T_{rh}$ for learning HBs and $T_{rk}$ learning the number of nearest neighbors HBs.

Step H2. Learn *dominant HBs*.

Step H3. Learn the number $k$ of nearest neighbors HBs.

Step H4. Generate rules $R_1$-$R_3$.

Step H5. Validate rules $R_1$-$R_3$ using validation data $V_d$.

Below we present Steps H2 and H3 in more details.

**Step H2**. Learn *dominant HBs*.

Step H2.1. Construct all pure HB on training data $T_r$ by the algorithm presented in Section II.B.

Step H2.2. Collect all n-D points {$b_j$} that belong to *single-point HBs* (HBs without other n-D points).

Step H2.3. For each $b_j$ find a nearest HB. If the class of $b_j$ and the nearest pure HB is the same, record it as a positive classification, else record it as negative classification.

Step H2.4. Attempt to combine HBs to larger dominant HBs by adding small HBs of other classes to adjacent large HBs keeping HB purity above in a predefines threshold *T*.

**Step H3**. Learn the number of nearest neighbors HBs *k*.

Step H3.1. Set up limits for the number *k* of nearest HBs.

Step H3.2. Loop through *k* is this interval to find *k* with most accurate voting classification for all n-D points in $T_{rk}$. If predefined accuracy threshold Q is reached, finish learning process, else leave n-D points in $T_{rk}$ unclassified and exit.

*A.2. Sets of hyper-blocks as generalized decision trees*

Below we discuss learning **simple dominant HBs** such as common in **decision trees**. According to the HB definition it satisfies inequalities $\| x_i - c_i \| \leq L_i / 2$ for respective *center* n-D point $\mathbf{c} = (c_1, c_2, \ldots, c_n)$ and *lengths* $\mathbf{L} = (L_1, L_2, \ldots, L_n)$.

Consider an example of a hyper-blocks $\| x_i - c_i \| \leq c_i$ in parallel coordinates that are in [0,10] interval. This inequality is true for all non-negative $x_i \leq 2c_i$, in other words, $x_i \in [0, 2c_i]$. Such simple HB needs only $c_i$ values that identify its center. Similarly, consider a complimentary HB where $x_i \in (2c_i, 10]$, i.e., $x_i > 2c_i$. Its center $\mathbf{h} = (h_1, h_2, \ldots, h_n)$ is defined by $h_i = (10 - 2c_i)/2$. As we see these HBs are defined by a simple set of inequalities.

The Hyper algorithm allows discovering and visualizing hyper-blocks of the type of branches of decision trees (DTs). It is based on a direct link between DTs and hyper-blocks. Consider, a branch of DT: $x_1 > 5$ & $x_2 < 6$ & $x_3 > 2$ that is labeled, say, by class 1. Assume that all coordinates $X_i$ are in [0.10] interval, then HB for this branch is defined by three intervals:

$x_1 \in (5, 10]$, $x_2 \in [0, 6)$ and $x_3 \in (2, 10]$.

Alternatively, instead of staring from a DT, we can start from a hyper-block. Let HB be given by three intervals $x_1 \in (5, 7)$, $x_2 \in (3, 6)$, and $x_3 \in (2, 4)$ within interval [0,10]. It is equivalent to the DT branch:

$x_1 > 5$ & $x_1 < 7$ & $x_2 > 6$ & $x_2 > 3$ & $x_3 > 2$ & $x_3 < 4$   (4)

These examples show a 1:1 mapping between a DT branch and a hyper-block without loss of generality.

A user can create a DT outside of VisCanvas 2.0 and then convert each DT branch to a respective HB to be visualized in VisCanvas 2.0. Alternatively, a set of HBs produced in VisCanvas 2.0 can be viewed as a set of DTs as follows.

Assume that each discovered HB is represented by a DT branch like (4). Can we combine these branches to a single tree? It is possible only in a special situation when branches have a common root, like in the following example. Say, branch *A* contains inequality $x_i \leq T$, and branch *B* contains the opposite inequality $x_i > T$, then coordinate $X_i$ can serve as a common root for *A* and *B*. This analysis shows that outside of special cases, a set of HBs (HB "*forest*") is a *more general model* than a DT model. A set of HBs removes a limitation of a single DT requiring a common root.

## B. WBC case study

### B.1. Learning of hyper-blocks with all data

WBC data have been used for exploration of the Hyper algorithm in comparison with ID3 decision tree algorithm implemented in VisCanvas 2.0 and Scikit-learns, respectively. First, we attempt to discover a hyper-block $HB_i$ that is dominant for class $C_i$ and then to build a rule on all WBC data:

If $\mathbf{x} \in HB_i \Rightarrow \mathbf{x} \in$ class $C_i$ dominant in $HB_i$   (5)

The use of all WBC data for training (discovering HBs) without setting aside a subset for validation allows getting the *best-case accuracy estimate* possible on WBC data with HBs.

Table 1 presents the HBs generated by the Hyper algorithm when using a 0.2 distance from the center of HB in WBC data normalized to [0,1]. Total 22 HBs have been produced: 20 pure and 2 mixed. Out of 20 pure HBs 7 belong to the class B (benign) and 13 to class M (malignant). Two mixed HBs are very different. One is almost pure class M dominant, 92/1, but the second HB is not dominant at all with 2 cases and ratio 1/1. In this situation when $HB_i$ is not dominant for any class, all n-D points of this HB are classified to class M (malignant) to avoid more dangerous misclassification to the benign class.

Table 1. Hyper-blocks with using a 0.2 center distance with the WBC dataset.

| Type of HCs | Number of HCs | Number of n-D points in HCs |
|---|---|---|
| Pure | 20 | 588 |
| Mixed | 2 | 95 |
| Total | 22 | 683 |

The number of n-D points in each HB also varies very significantly from 404 n-D points to a single n-D point for class B, and from 76 n-D points to single n-D point for class M (see Table 2). The produced HBs overlap due to the version of the algorithm used. A more elaborated version allows removing overlaps. Having only two mixed HBs with ratios 92/1 and 1/1 the Hyper algorithm misclassified only two benign cases with total accuracy 681/683, i.e., 99.70%.

Table 2. Number of n-D points in pure overlapped HBs for WBC dataset.

| class | Number of n-D points in hyper-block | | | | | | | | | | |
|---|---|---|---|---|---|---|---|---|---|---|---|
| B | 404 | 392 | 92 | 34 | 16 | 5 | 4 | 1 | | | |
| M | 76 | 44 | 42 | 39 | 34 | 25 | 18 | 18 | 12 | 9 | 7 | 1 |

This rule relies on several small HB, where 7 HBs contain 9 or less n-D points, which can be interpreted as overfitting. One of the ways to avoid it is refusing to classify n-D points that are in these HBs, i.e., removing them from the rule. These HBs contain 29 cases. It decreases the number of cases covered by the rule to 683-29=654 (recall 95,75%) with precision 99.85% (653/654) that is slightly greater than the accuracy 99.7%. The HB threshold increase to 25 cases removes 11 HBs with 91 n-D points, resulted in recall of 86.68% and precision of 99.83. These precisions set up two other *best-case benchmarks* for HB-based algorithms on WBC data.

### B.2. Supervised learning with training and validation data

**10-fold cross validation** was performed on the WBC dataset using the VisCanvas 2.0 and Scikit-learns ID3 decision tree implementation. The Hyper algorithm in VisCanvas 2.0 automatically created hyper-blocks for each 10-fold split, then rules (1)-(3) formed from these HBs have been tested on the validation folds. This includes using a k-NN approach for n-D points that do not belong to any HB. Three versions of *k*-nearest neighbors were explored for each n-D point **x** with *k*=1, 3, 5:

(N1) the distance from **x** to the *center n-D points* of discovered hyper-blocks,

(N2) the distance from **x** to the *mean n-D point* of discovered hyper-blocks,

(N3) the distance from **x** to the *nearest n-D point*

For N1-N3 with pure HBs, Table 3 presents the results that are similar to each other with average accuracy above 95% and reaching 97.61%. N3 produced the lowest results indicating that it is more sensitive to the distribution of closest points, while N1 and N3 based on means and centers are more robust. The average accuracy for the ID3 decision tree is 92.85%, with min equal to 89.86% and max equal to 94.58% that are **below** the averages for N1-N3 showing the advantages of HBs for this dataset. The DT has depth 8 and 23 branches which is greater than the average number of HBs which is 18.1 in the worst case.

Likewise, for N1-N3 with mixed HBs, Table 3 presents the results with similar accuracy scores. This could be because the number of HBs has been reduced to approximately 6 in comparison with the pure HBs average of 18. These HBs cover more space and it is easier to place n-D points within them for classification. This means that the k-NN version of the algorithms will be rarely needed, and so accuracies would be consistent and there would be no need for N1, N2, or N3. It also stayed above the averages of the decision tree except in the case of minimum accuracies where they were lower than the decision tree. Table 5 demonstrates the comparable ID3 decision tree on training data. Table 6 shows a confusion matrix of this in ID3 decision trees on the representative validation fold where the number of errors is greater for the cancer class than produced by the Hyper algorithm.

Table 3. Summary of experiments with WBC data using pure and mixed HBs.

| Model type | 10-fold accuracy | | | average # of HB |
|---|---|---|---|---|
| | average | min | max | |
| *k*=1 | | | | |
| N1 pure | **95.52** | 91.04 | 100 | 16.9 |
| N2 pure | 95.22 | 89.55 | 100 | 16.9 |
| N3 pure | 93.58 | 86.57 | 100 | 16.9 |
| N1 mix | 93.43 | 86.57 | 97.01 | 5.4 |
| N2 mix | **93.58** | 86.57 | 97.01 | 5.4 |
| N3 mix | 93.28 | 86.57 | 98.51 | 5.4 |
| *k*=3 | | | | |
| N1 pure | 96.57 | 92.54 | 98.51 | 17.3 |
| N2 pure | **97.61** | 94.03 | 100 | 17.3 |
| N3 pure | 94.78 | 91.04 | 97.01 | 17.3 |
| N1 mix | **93.88** | 88.06 | 98.51 | 7.1 |
| N2 mix | **93.88** | 88.06 | 98.51 | 7.1 |
| N3 mix | 93.73 | 88.06 | 97.01 | 7.1 |
| *k*=5 | | | | |
| N1 pure | **96.42** | 92.54 | 100 | 18.1 |
| N2 pure | **96.42** | 92.54 | 100 | 18.1 |
| N3 pure | 94.78 | 89.55 | 100 | 18.1 |
| N1 mix | **96.12** | 91.04 | 98.51 | 6.4 |
| N2 mix | **96.12** | 91.04 | 98.51 | 6.4 |
| N3 mix | 95.82 | 91.04 | 98.51 | 6.4 |

Table 4 shows confusion matrixes for the fold that is closest to the average accuracy in 10-fold cross validation for pure and mixed hyper-blocks used in the Hyper algorithm for *k*=1,3,5.

Table 4. Confusion matrixes for the fold that is closest to the average accuracy.

| | | | | | | | | | | |
|---|---|---|---|---|---|---|---|---|---|---|
| **Pure hyper-blocks** | | | | | | | | | | |
| *k*=1 | N1 | B | M | | N2 | B | M | | N3 | B | M |
| | B | 44 | 0 | | B | 44 | 0 | | B | 44 | 0 |
| | M | 3 | 20 | | M | 3 | 20 | | M | 4 | 19 |
| *k*=3 | N1 | B | M | | N2 | B | M | | N3 | B | M |
| | B | 43 | 1 | | B | 43 | 1 | | B | 44 | 0 |
| | M | 1 | 22 | | M | 1 | 22 | | M | 3 | 20 |
| *k*=5 | N1 | B | M | | N2 | B | M | | N3 | B | M |
| | B | 42 | 2 | | B | 43 | 1 | | B | 44 | 0 |
| | M | 0 | 23 | | M | 2 | 21 | | M | 3 | 20 |
| **Mixed dominant hyper-blocks** | | | | | | | | | | |
| *k*=1 | N1 | B | M | | N2 | B | M | | N3 | B | M |
| | B | 44 | 0 | | B | 42 | 2 | | B | 42 | 2 |
| | M | 3 | 20 | | M | 2 | 21 | | M | 3 | 20 |
| *k*=3 | N1 | B | M | | N2 | B | M | | N3 | B | M |
| | B | 43 | 1 | | B | 43 | 1 | | B | 44 | 0 |
| | M | 1 | 22 | | M | 1 | 22 | | M | 3 | 20 |
| *k*=5 | N1 | B | M | | N2 | B | M | | N3 | B | M |
| | B | 41 | 3 | | B | 43 | 1 | | B | 43 | 1 |
| | M | 0 | 23 | | M | 2 | 21 | | M | 3 | 21 |

Table 5. Number of n-D points in nodes of ID3 for WBC dataset.

| Cases | Number of n-D points of two classes in the node | | | | | | | | |
|---|---|---|---|---|---|---|---|---|---|
| Node | 1 | 2 | 3 | 4 | 5 | 6 | 7 | 8 | 9 |
| Class B | 397 | 364 | 33 | 363 | 1 | 15 | 18 | 363 | 0 |
| Class M | 217 | 9 | 208 | 2 | 7 | 3 | 205 | 1 | 1 |
| Node | 10 | 11 | 12 | 13 | 14 | 15 | 16 | 17 | 18 |
| Class B | 1 | 0 | 15 | 0 | 15 | 3 | 362 | 1 | 9 |
| Class M | 0 | 7 | 0 | 3 | 149 | 156 | 0 | 1 | 4 |
| Node | 19 | 20 | 21 | 22 | 23 | 24 | 25 | 26 | 27 |
| Class B | 6 | 2 | 1 | 1 | 0 | 8 | 1 | 5 | 1 |
| Class M | 45 | 9 | 147 | 0 | 1 | 0 | 4 | 16 | 29 |
| Node | 28 | 29 | 30 | 31 | 32 | 33 | 34 | 35 | 36 |
| Class B | 2 | 0 | 1 | 0 | 0 | 1 | 4 | 1 | 0 |
| Class M | 2 | 7 | 24 | 123 | 4 | 0 | 16 | 0 | 22 |
| Node | 37 | 38 | 39 | 40 | 41 | 42 | 43 | 44 | 45 |
| Class B | 1 | 2 | 0 | 0 | 1 | 3 | 1 | 1 | 0 |
| Class M | 7 | 0 | 2 | 20 | 4 | 4 | 12 | 1 | 6 |
| Node | 46 | 47 | 48 | 49 | 50 | 51 | 52 | 53 | 54 |
| Class B | 1 | 0 | 3 | 0 | 1 | 0 | 0 | 1 | 0 |
| Class M | 0 | 4 | 1 | 3 | 1 | 11 | 1 | 0 | 1 |
| Node | 55 | 56 | 57 | | | | | | |
| Class B | 3 | 0 | 1 | | | | | | |
| Class M | 0 | 1 | 0 | | | | | | |

Table 6. ID3 Confusion matrix.

| ID3 | 2 | 4 |
|---|---|---|
| B | 45 | 2 |
| M | 5 | 17 |

We compared the complexity of HBs vs. DT by estimating the amount of numbers needed to store HBs and DT. It had shown that this DT needs more numbers with ratio above 1.5.

*B.3. Merger*

The steps of the Merger Hyper-blocks (**MHyper**) algorithm (see section II.B) were performed on the WBC dataset in conjunction with 10-fold cross validation. The average number of HBs created for each fold was 17.3. Four of the folds contain HBs that include only a single n-D point. Three folds had one HB with only a single n-D point and another fold had two HBs with only a single n-D point. To help this approach gain more generality it was adapted to also consider **dominant** HBs. After performing the previously outlined steps, we attempted to reduce the number of single points HBs including impure combinations. All steps of MHyper algorithm were performed with a threshold of 10% impurity. It produced 5 HBs in average for each fold without any single n-D point HB. This approach of defining a threshold for the percentage of n-D points from an opposite class allowed a balance between generalization and accuracy to be made. Table 7 shows the resulting 5 HBs in a representative fold. The merger algorithm allows overlap of merged HBs which is seen in Table 7. The average number of HBs decreases about 3 times from the initial number of pure HBs. All these HBs have impurity less than 10% (see Table 7).

Table 7. Results of MHyper for impure HBs for a representative fold.

| Hyper-block | Number of n-D points of two classes in HBs | | | | |
|---|---|---|---|---|---|
|  | 1 | 2 | 3 | 4 | 5 |
| Cases of Class B | 25 | 0 | 439 | 432 | 12 |
| Cases of Class M | 226 | 53 | 47 | 30 | 137 |
| impurity | 9.9% | 0 | 9% | 6.5% | 8% |

We also compared complexity of HBs vs. the decision tree by using the number of cases in the smallest merged HB. This merged HB contains 53 cases, which is 7.76% of the overall data. The DT's minimum number of n-D points within its nodes is 1. This is 0.15% of the overall data. It has 10 of these nodes with a single element within it. It also has 46 nodes that contain less than 53 cases. So, 80.7% of the DT's nodes are less generalized than the HB's least generalized block. The merger of HBs and pruning of DTs leads to higher generalization and higher error rate that is controlled by the impurity threshold.

*B.4. Simple hyper-boxes and dimension reduction*

Below we present experiments for rule discovery by the Hyper algorithm with simple HBs of the type of decision tree branches. A simplest discovered rule in WBC data is a **one-dimensional rule**

     if $x_6 < 3$ then class benign else class malignant     (6)

This rule has accuracy 91.22% with correctly classified 623 cases out of 683 total cases. A **two-dimensional rule** is

    if $x_6<3$ & $x_8<4$ then class B (benign) else class M    (7)

This rule has accuracy 93.85% with correctly classified 641 cases out of 683 total cases. A **three-dimensional rule** is

if $x_6<3$ & $x_8<4$ & $x_5 < 6$ then class 2 (benign) else class M  (8)

This rule has accuracy 94.58% with correctly classified 646 cases out of 683 total cases. The rules have been discovered in a dimension reduction process in VisCanvas 2.0 where a user can toggle coordinates on and off in the "Coordinate" tab of the settings to reduce data dimension. This is resulted in reduction of WBC dataset from 9-D down to the 3-D where patterns were perceived previously visually. Fig. 14 displays two clearly separated HBs of opposite classes where the violet HB contains 95 malignant cases, and the yellow HB contains 404 benign cases. This is a large majority of the cases and the mirror like difference of them makes the pattern easier to distinguish.

V. CONCLUSION

This paper contributes to interpretable machine learning via visual knowledge discovery in parallel coordinates by visual pattern discovery, data and classification model visualization, dimensionality reduction, and model simplification. It allows putting the end-user in the driver seat of model development.

The concepts of *hypercubes* and *hyper-blocks* are used as the major concepts in this study to allow end-users to easily understand patterns. The proposed Hyper algorithm was tested on WBC data. It is shown that Hyper models generalize decision trees.

Major benefits of the Hyper algorithm are the abilities to discover and observe HBs by end-users including observing all HBs together side-by-side like in Fig. 12. It allows detailed visual analysis of individual HBs of each class. HBs are well understood by the end-users being directly presented in the original attributes with the ability to cut down unimportant attributes. Another advantage relative to the decision trees is extended abilities to avoid data overgeneralization.

The paper also outlined VisCanvas 2.0 software that implements the HB approach. Future studies will include maturing the Hyper algorithm and software to larger datasets with GPU and multithreading.